\documentclass[lettersize,journal]{IEEEtran}
\usepackage{amsmath,amsfonts}
\usepackage{algorithmic}
\usepackage{algorithm}
\usepackage{array}
\usepackage{textcomp}
\usepackage{stfloats}
\usepackage{url}
\usepackage{verbatim}
\usepackage{graphicx}
\usepackage{cite}
\usepackage{tabularray}
\usepackage{multirow, subcaption, colortbl}
\usepackage{hyperref}

\begin{document}

\title{BlanketGen2-Fit3D: Synthetic Blanket Augmentation Towards Improving Real-World In-Bed Blanket Occluded Human Pose Estimation}

\author{Tamás Karácsony, João Carmona, João Paulo Silva Cunha~\IEEEmembership{Senior Member,~IEEE,}
\thanks{This work was partially funded by Fundação para a Ciência e a Tecnologia under the scope of the CMU Portugal program Ref PRT/BD/152202/2021. DOI 10.54499/PRT/BD/152202/2021 (https://doi.org/10.54499/PRT/ BD/152202/2021). This work is financed by National Funds through the Portuguese funding agency, FCT - Fundação para a Ciência e a Tecnologia, within project LA/P/0063/2020. DOI 10.54499/LA/P/0063/2020 | https://doi.org/10.54499/LA/P/0063/2020}
}

\markboth{Journal of \LaTeX\ Class Files,~Vol.~14, No.~8, August~2021}%
{Shell \MakeLowercase{\textit{et al.}}: A Sample Article Using IEEEtran.cls for IEEE Journals}


\maketitle

\begin{abstract}
Human Pose Estimation (HPE) from monocular RGB images is crucial for clinical in-bed skeleton-based action recognition, however, it poses unique challenges for HPE models due to the frequent presence of blankets occluding the person, while labeled HPE data in this scenario is scarce.
To address this we introduce BlanketGen2-Fit3D (BG2-Fit3D), an augmentation of Fit3D dataset that contains 1,217,312 frames with synthetic photo-realistic blankets. To generate it we used BlanketGen2, our new and improved version of our BlanketGen pipeline that simulates synthetic blankets using ground-truth Skinned Multi-Person Linear model (SMPL) meshes and then renders them as transparent images that can be layered on top of the original frames. 
This dataset was used in combination with the original Fit3D to finetune the ViTPose-B HPE model, to evaluate synthetic blanket augmentation effectiveness. The trained models were further evaluated on a real-world blanket occluded in-bed HPE dataset (SLP dataset). 
Comparing architectures trained on only Fit3D with the ones trained with our synthetic blanket augmentation the later improved pose estimation performance on BG2-Fit3D, the synthetic blanket occluded dataset significantly to (0.977 Percentage of Correct Key-
points (PCK), 0.149 Normalized Mean Error (NME)) with an absolute 4.4\% PCK increase. Furthermore, the test results on SLP demonstrated the utility of synthetic data augmentation by improving performance by an absolute 2.3\% PCK, on real-world images with the poses occluded by real blankets. These results show synthetic blanket augmentation has the potential to improve in-bed blanket occluded HPE from RGB images.

The dataset as well as the code will be made available to the public upon acceptance.
\end{abstract}


\begin{figure*}
    \includegraphics[width=1\textwidth]{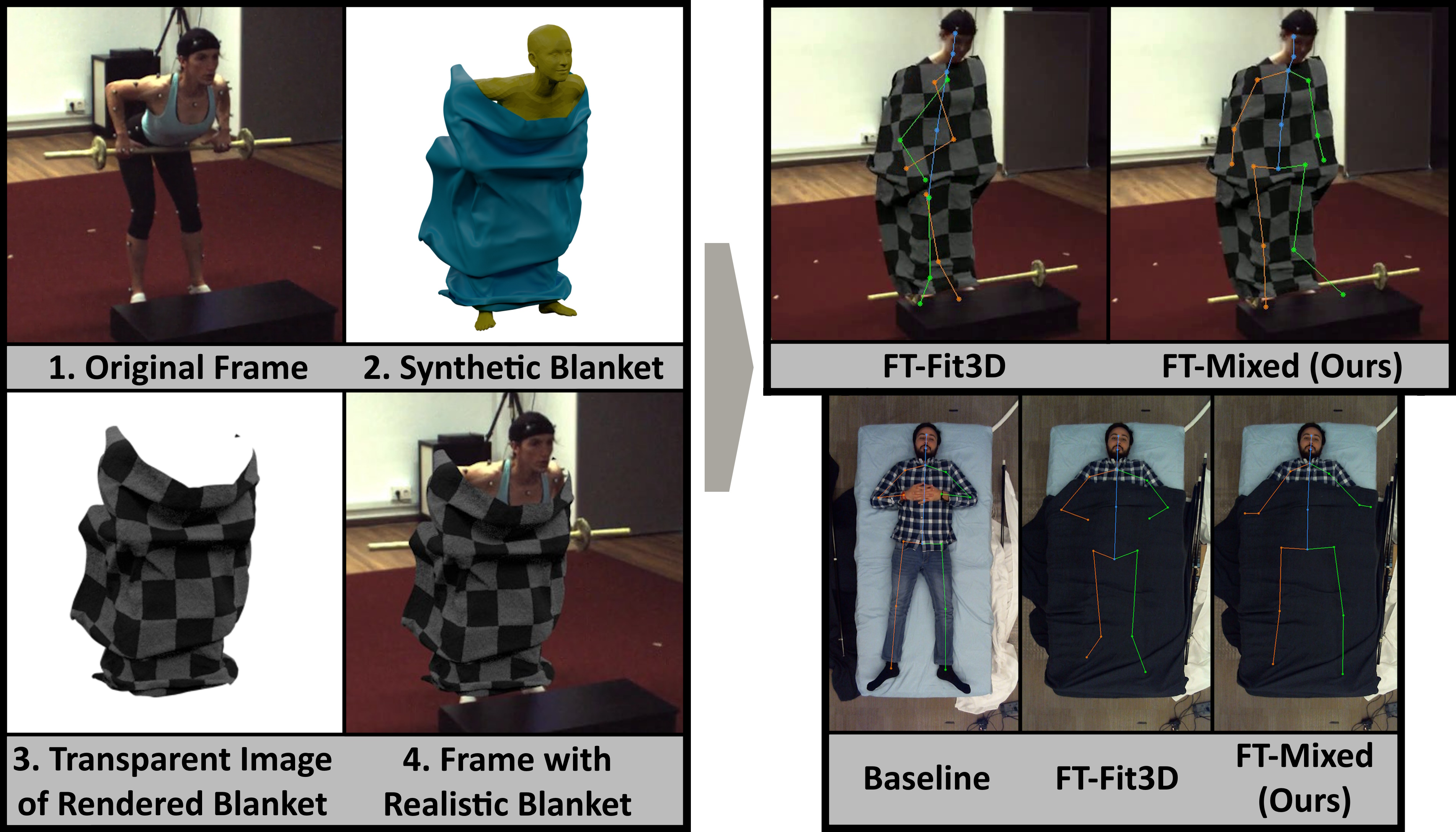}
    \caption{On the left: different stages of the {BlanketGen2} pipeline with the improvements introduced in this paper. On the right: pose predictions on a frame from {BlanketGen2-Fit3D} and another from SLP with two different fine-tuned ViTPose-B models \cite{vitpose}, one fine-tuned on Fit3D (FT-Fit3D) and the other on a mixed dataset of Fit3D and {BlanketGen2-Fit3D} (FT-Mixed).}
    \label{fig:abstract}
\end{figure*}

\section{Introduction}
\label{sec:intro}
\IEEEPARstart{H}{uman} pose estimation from monocular images represents a significant challenge in computer vision. This difficulty stems from the inherent ambiguity of data and the vast range of visual contexts in which it is applied. Despite these challenges, recent advancements have achieved remarkable results. By leveraging large datasets, these methods train deep-learning models that are increasingly adept at understanding and interpreting complex human poses. \cite{vitpose,hybrik,rtmpose,CLIFF}. 

However, it is nearly impossible to generalize beyond contexts with available labeled data using these methods. One such context is in-bed patient monitoring: the presence of blankets prevents the use of optical systems for acquiring ground truth poses, and inertial measurement units are uncomfortable to wear and tend to move around due to regular contact with the bed which reduces their accuracy, so available datasets are rare and usually not large enough to train big architectures \cite{tamasreview}.

Synthetic data generation provides a potential solution to this problem. Previous works \cite{bedlam,AGORA} have demonstrated the potential of this type of approach in broader scenarios, so it only makes sense to attempt it in more specific contexts with more targeted data synthesis. In \cite{blanketgen} we implemented a pipeline to augment the 3DPW dataset \cite{3DPW} with synthetic blankets and the augmented dataset (BlanketGen-3DPW) was used to fine-tune a HybrIK \cite{hybrik} model with promising results on the simulated data. 

Here we present a new version of our BlanketGen pipeline and employ it to augment the significantly larger Fit3D dataset \cite{fit3d} with realistic synthetic blankets; then utilize the resulting dataset, BlanketGen2-Fit3D, to train the ViTPose-B HPE model \cite{vitpose} and test this new model on the SLP dataset \cite{slp} which contains real-world images of people posing in bed with real blankets over them (14,715 frames). We show that synthetic photo-realistic blanket augmentation is a promising method to improve real-world in-bed 2D human pose estimation.

From the implementation aspect, we improved BlanketGen by separating the rendering and cloth simulation, this allows the rendering of different blankets without having to rerun the cloth simulations, with the added bonus of using both GPU processing power and CPU processing power more efficiently since the cloth simulation runs only on CPU whereas the rendering mostly runs on GPU. Additionally, we also improved the textures of the blankets to be more realistic. Fig.\ref{fig:abstract} showcases different steps of the pipeline as well as pose estimations with two different models. 

BlanketGen was originally developed to augment 3DPW, however, it is a rather small dataset (53,796 frames), therefore we chose Fit3D instead as it contains significantly more labeled frames (1,782,318 frames) and its interior setting makes it more similar to real scenarios where in-bed pose estimation is useful. The consistent lighting is also easier to replicate in the rendering which allows for more realistic blankets.

The choice to use a 3D dataset for 2D pose estimation was necessitated by the constraint that the original dataset must have ground truth body meshes of the participants to enable the cloth simulation, as well as the only available dataset with real blankets and ground truth pose estimations being SLP, which is a 2D dataset. 


\textbf{Our main contributions are:}

\begin{itemize}
    \item \textbf{BlanketGen2}: A new version of BlanketGen with photo-realistic blanket textures and an efficient two step cloth simulation and rendering, allowing multiple rendering of scenes without rerunning the cloth simulation, enabling online augmentation with new blanket textures while training;
    \item \textbf{BlanketGen2-Fit3D}: A \textit{new dataset} consisting of Fit3D with synthetic blanket augmentations, including both the rendered blanket frames and the .blend files that allow for rendering more frames;
    \item A benchmark for pose estimation on BlanketGen2-Fit3D with a \textbf{ViTPose-B model fine-tuned} on the mix of Fit3D and BlanketGen2-Fit3D (0.977 PCK, and 0.149 NME, Table \ref{tab:results});
    \item Tests of the performance on real blankets on the SLP dataset, that showed training on \textbf{BlanketGen2-Fit3D improves results with real blankets} (PCK improved by 0.023, Table \ref{tab:results});
\end{itemize}

\section{Related Works}
\label{sec:formatting}
There are numerous datasets with real-world and in-lab images and videos for pose estimation, some providing 3D ground truth poses, often with multiple camera viewpoints and body meshes as well (e.g. 3DPW, Human3.6m, Fit3D) \cite{ikea,behave,huMMan,h36m,fit3d,3DPW,aist++,humansc3d,chi3d,humaneva}, others contain only 2D pose labels (e.g. MS-COCO, MPII) \cite{mscoco,mpii,lsp,crowdpose,OCHuman}, some with body segmentation ground truth too.

Existing fully synthetic datasets provide almost perfect ground truth such as AGORA or Surreal \cite{AGORA,bedlam,Deep3DPose,gtahuman,surreal}, however, these are generally not very realistic since generating realistic-looking humans is extremely difficult.

Augmentation of real images with synthetic data has also been explored \cite{3dhp,muco,3dhumans}, in particular, augmentation of depth videos with synthetic blankets has received some attention. In \cite{patientMoCap} depth videos of people performing specified movements in a hospital bed were augmented with synthetic blankets and the resulting dataset was used for training and testing 3D pose estimation models; still, models trained with it were not tested with real blankets. Similarly, in \cite{ochi2022} depth images were augmented with synthetic blankets but were tested on data with real blankets, with training on a mixed dataset of real and synthetic data providing the best results. To estimate body pose and contact pressure on real beds with real blankets from depth images \cite{clever2021} generated a fully synthetic dataset with good results. All of these approaches used depth images, therefore they could not very effectively leverage the available large datasets and research advancements in pose estimation from RGB images. In \cite{blanketgen}, where the BlanketGen pipeline was first introduced, a dataset (BlanketGen-3DPW) with real images was augmented with synthetic blankets and used for 3D pose estimation, however, the blankets were not very realistic and the dataset size was relatively limited.

Datasets with real blankets are scarce, the SLP dataset was introduced in \cite{slp} and provides RGB, depth, and Long-wave Infrared images as well as pressure mat data of people lying in bed holding different poses. Every pose was recorded with no blanket occlusions and with 2 different blankets with the corresponding 2D body pose ground truths. This approach presents an intriguing method for addressing the challenge of providing keypoint labels under blanket occlusion. However, since the blankets are placed over participants who maintain a static pose, this method fails to capture the dynamics and deformation that occur during actual movements in contact with the blanket. Moreover, the size of the dataset is relatively small. The only other dataset available related to this domain, to the best of the author's knowledge, is BlanketSet \cite{blanketset}, which is an in-bed blanket occluded action recognition dataset. While it does not contain keypoint labels, it can still be utilized for the qualitative evaluation of models, such as BlanketGen-3DPW presented in \cite{blanketgen}.

\section{Methods}

\subsection{Datasets}
\subsubsection{Fit3D dataset}
We choose Fit3D \cite{fit3d} for the augmentation as it has ground-truth SMPL body meshes, joint locations, and is a relatively large dataset (1,782,318 frames) for the currently available 8 subjects. Another big advantage of Fit3D for this application is that the movements are confined to a small area, this is beneficial as simulated blankets generally do not look very realistic on people who are moving from one location to another, and tend to fall off pretty quickly. Furthermore, Fit3D's four cameras allow for the generation of four frames, from the four different viewpoints, for every frame of the blanket simulation.

The movement sequences were split into three categories: standing up, lying down, and mixed. On standing-up sequences the gravity direction used in the cloth simulation was set as perpendicular to the torso at the starting position, on lying-down sequences it was set pointing downwards perpendicular to the floor. Mixed sequences (where the subject alternated between standing and lying down or sitting) were excluded from BlanketGen2-Fit3D since the blanket simulations did not look realistic (Tab. \ref{tab:blanketgen3_fit3d}).

\begin{table}[htbp]
  \centering
  \resizebox{0.48\textwidth}{!}{\begin{tabular}{|l|c|c|c|c|}
    \hline
     & \textbf{Standing} & \textbf{Lying} & \textbf{Alternating} & \textbf{Total}\\
    \hline
    \textbf{Subjects} & \multicolumn{4}{c|}{8} \\
    \hline
    \textbf{Exercise Types} & 36 & 2 & 9 & 38\\
    \hline
    \textbf{Videos} & 628 & 27 & 0 & 655\\
    \hline
    \textbf{Frames} & 1,190,880 & 26,432 & 0 & 1,217,312\\
    \hline
    \textbf{Resolution} & \multicolumn{4}{c|}{900x900}\\
    \hline\arrayrulecolor{black}
    \textbf{Framerate} & \multicolumn{4}{c|}{50 fps} \\
    \hline
  \end{tabular}}
  \caption{Information about the generated BlanketGen2-Fit3D dataset. Note that all the alternating sequences present in Fit3D were unused therefore they are not counted for the total. Standing, lying, and mixed refer to whether the sequences were performed standing up, lying down, or alternating between the two.}
  \label{tab:blanketgen3_fit3d}
\end{table}

\subsubsection{SLP dataset}
SLP was chosen for testing since it contains real blankets as well as ground-truth position labels. 

It consists of 14,715 frames in total, from that 4,905 pictures of people lying in beds without blankets covering them as well as two pictures with different blankets for every one of the pictures without blankets for a total of 9810 pictures with blankets. It also contains different modalities of data than RGB images, but for this work they were not utilized.

It was used only for testing, thus there was no need to split it into train, validation, and test sets, instead it was split into pictures with blankets and without blankets to evaluate the performance of the models in both of these scenarios separately.

\subsection{Synthetic blanket generation: BlanketGen2-Fit3D dataset}

\begin{figure}[htb]
\centering
\includegraphics[width=1\linewidth]{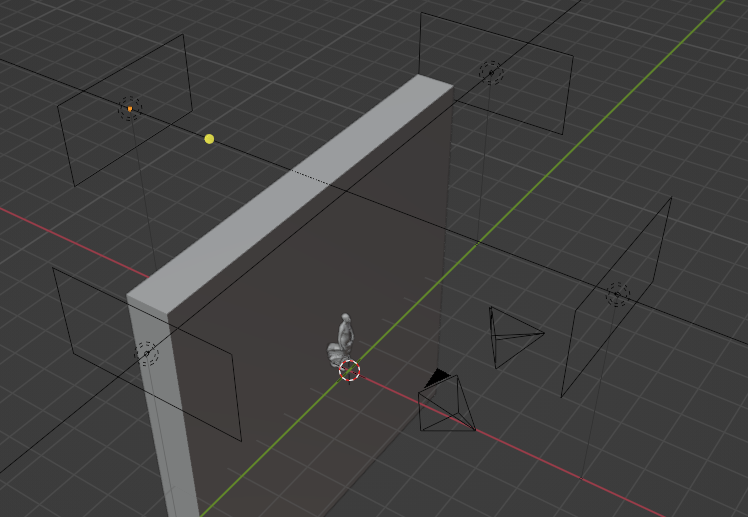} 
\caption{The scene from BlanketGen2-Fit3D, the four big rectangles are area lights set up to mimic the light reflecting off the walls and ceiling in the room; the rectangular cuboid behind the person serves the purpose of a bed for the physics of the blanket and is completely excluded from the rendering; the pyramids are representing the cameras.}
\label{fig:scene}
\end{figure}

The synthetic blankets are simulated for the Fit3D dataset in Blender \cite{blender}. Here, a physics-based cloth simulation engine models the fabric dynamics, where the simulated blanket interacts with the SMPL mesh of the person, additionally accounting for realistic collisions with a simulated bed surface positioned behind them. (Fig. \ref{fig:scene}). The simulated scene is designed to replicate the original recording environment, matching lighting conditions and camera angles. Once simulated, the blankets are textured and rendered as images with a transparent background. These rendered blankets are then seamlessly composited onto the original frames, enabling their independent distribution without the licensing constraints associated with the original dataset (Fig.\ref{fig:pipeline_new}). 

The simulated and rendered blanket dataset (BlanketGen2-Fit3D) will be made available to the public.

\subsubsection{Efficiency of cloth simulation and rendering in two steps}

\begin{figure*}[tb]
\centering
\begin{subfigure}[t]{0.45\linewidth}
\includegraphics[width=0.99\linewidth]{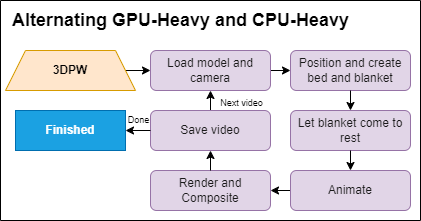} 
\caption{Diagram of the old BlanketGen pipeline as described in \cite{blanketgen}.}
\label{fig:pipeline_old}
\end{subfigure}
\hspace{0.05\linewidth}
\begin{subfigure}[t]{0.45\linewidth}
\includegraphics[width=0.99\linewidth]{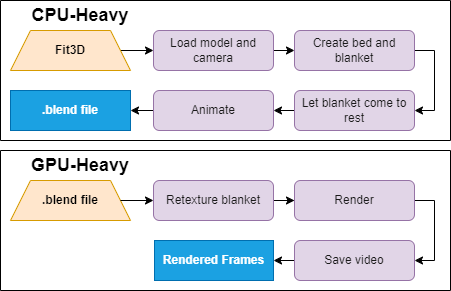} 
\caption{Diagram of the new BlanketGen2 pipeline presented in this paper.}
\label{fig:pipeline_new}
\end{subfigure}
\caption{Comparison of the BlanketGen and BlanketGen2 data processing pipelines, highlighting the enhanced efficiency and versatility of the new setup (b). The BlanketGen2 pipeline also introduces the capability for unrestricted sharing of generated blankets, independent of the source dataset, and supports flexible retexturing options.}
\end{figure*}

The synthetic blanket generation pipeline is split into two parts: the loading, cloth simulation, and animation in the first one, and the rendering in the second one  (Fig.\ref{fig:pipeline_new}). The main advantage of this split is that the baked cloth simulations are all saved, in a .blend file, which allows for rendering with different parameters, the 4 viewpoints, and textures without having to rerun all of the cloth simulations. Furthermore, as the first part runs exclusively on the CPU whereas the second one runs mostly on the GPU, separating them allowed the pipeline to be more easily adapted to changing hardware availability by using as much CPU and GPU processing power as is available at any given moment. This significantly sped up processing time compared to the old BlanketGen pipeline which followed a single 6-step process to generate the augmented videos (Fig.\ref{fig:pipeline_old})

\begin{figure}[tb]
\begin{subfigure}[t]{.49\linewidth}
  \centering
  \includegraphics[height=24em]{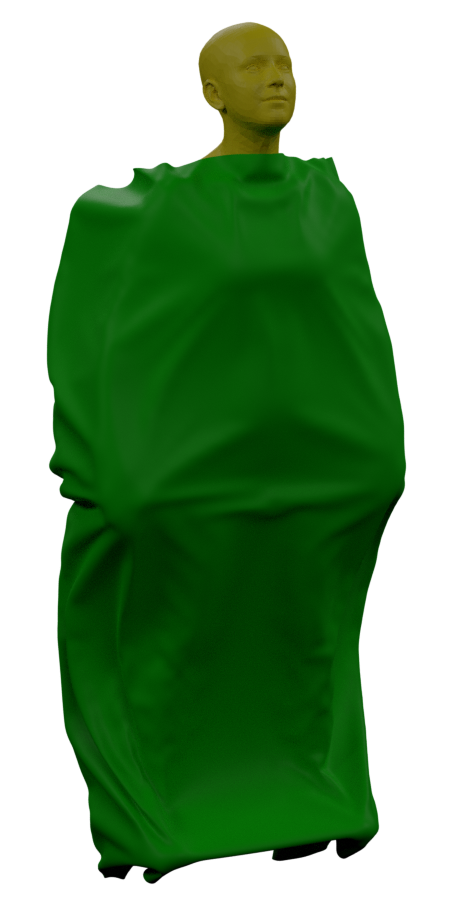} 
  \label{fig:ogblanket}
\end{subfigure}
\begin{subfigure}[t]{.49\linewidth}
  \centering
  \includegraphics[height=24em]{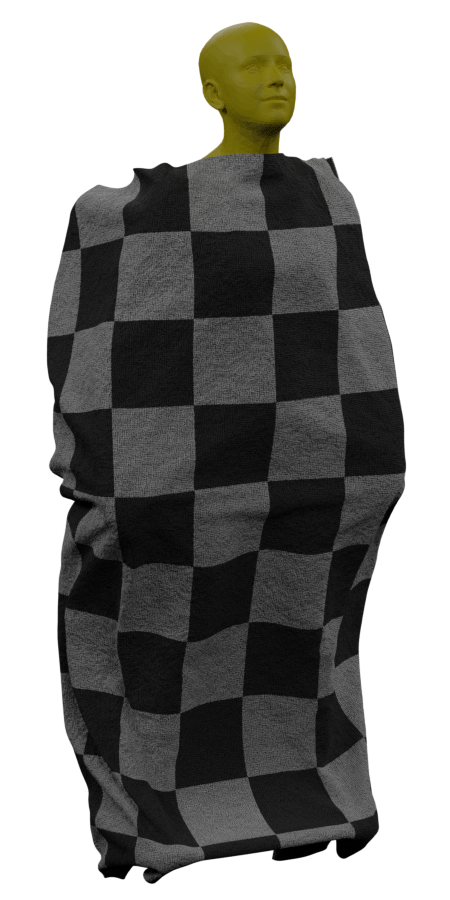} 
  \label{fig:newblanket}
\end{subfigure}
\setlength{\belowcaptionskip}{-10pt}
\caption{The blanket texture used in the old BlanketGen pipeline on the left, and on the right the new, realistic texture used for BlanketGen2-Fit3D.}
\label{fig:blankets}
\end{figure}

\subsubsection{Photo-realistic blanket textures}

In order to imitate realistic blanket textures we added two perpendicular distorted sinusoidal patterns to simulate the pattern of a woven fabric, with bump mapping to add the perception of depth to the texture. A grey and black checkerboard pattern was also used to make it less uniform (Fig.\ref{fig:blankets}). This approach significantly enhanced the visual realism of the blanket, achieving a more lifelike appearance, especially in comparison to the previous version (Fig.\ref{fig:blankets})

The interior setting of Fit3D with 4 fixed cameras and indoor lighting allowed for a more realistic replication of lighting conditions in the rendering, thus lights were manually placed to match the ones in the room where the dataset was recorded (Fig.\ref{fig:scene}).

\subsection{ViTPose Human Pose Estimation Architecture}

For our experiments, we selected the pre-trained ViTPose-B model for human pose estimation (HPE) \cite{vitpose}. ViTPose is a Vision Transformer-based architecture that leverages extensive pretraining on multiple datasets, offering scalability to accommodate different levels of computational resources. One key advantage of this architecture is its fine-tuning flexibility, enabling the model to generalize effectively to various tasks by adjusting only a subset of parameters \cite{vitpose}. This adaptability allows efficient optimization for specific use cases.

Additionally, ViTPose incorporates inherent occlusion-aware pretraining through the Masked Autoencoder (MAE) approach \cite{he2022masked}. During MAE pretraining, 75\% of image patches are randomly masked, and the model is trained to reconstruct the missing information using a combined dataset of MS COCO \cite{mscoco} and AI Challenger \cite{aic}. This enhances the model's robustness to occlusions, which is crucial in our simulated environment where the blanket interacts with the person and obscures key body parts.

In all experiments, we utilized the ViTPose-B model \cite{vitpose} with an input resolution of 256x192 pixels, consisting of 86 million parameters, implemented in the MMPose codebase \cite{mmpose}.

\subsection{Experiments}

We fine-trained and evaluated in multiple settings the ViTPose-B model \cite{vitpose}, utilizing the Fit3D, and the generated Blanketgen2-Fit3D datasets, with subjects S3, S4, S5, S7, S8, S9 being used for the train set, S10 for the test set, and S11 for the validation set. Additionally, we evaluated the trained architectures on the SLP dataset to assess their effectiveness in real-world scenarios involving blanket occlusions, demonstrating the practical utility of our generated dataset.

In all experiments, we used ground truth bounding boxes for the subjects, calculated by identifying the outermost ground truth joints in each direction and extending the boundaries by an additional 30 pixels.

\subsubsection{VitPose training and evaluation on Fit3D and Blanketgen2-Fit3D}

First, we trained two deconvolutional heads for the VitPose-B model (FT-Fit3D and FT-Mixed). 
As Fit3D and BlanketGen2-Fit3D datasets are less diverse than the datasets utilized for the original VitPose training we opted first for only fine-training the heads to avoid overfitting to this dataset and conserve the advantages of the pre-trained architecture. 
Moreover, as these datasets utilize an extension of the skeleton format introduced in Human3.6M \cite{h36m}, while the ViTPose model was pre-trained on datasets using the MS-COCO format \cite{mscoco}, we fine-trained a head specifically for Fit3D as a baseline (FT-Fit3D), using only the Fit3D dataset.
By fine-tuning only the deconvolutional head we could evaluate whether the features extracted by the VitPose backbone provide sufficient information to adjust to this new joint format. 

To evaluate the utility of using synthetic blanket occlusion augmented data (Blanketgen2-Fit3D) for training the HPE architecture we fine-trained the head (FT-Mixed) on a combined dataset of the original Fit3D and Blanketgen2-Fit3D. By training on the mix of these two datasets, we aim to enhance generalization across occluded and non-occluded scenarios, avoiding overfitting to only occluded scenarios.

In both models (FT-Fit3D, FT-Mixed), the backbone was frozen and we trained the head with a batch size of 128, a learning rate of 5e-5 for up to 20 epochs using the mean per joint position error on the validation set for early stopping, with validation being run every 2 epochs. Additionally, we experimented with unfreezing the backbone as well as the head and further fine-tuning the full model with the same settings except with a reduced learning rate of 1e-5 producing BB-Fit3D and BB-Mixed respectively.

This training setup produced four distinct model architectures: two with the pre-trained backbone and only the heads trained (FT-Fit3D and FT-Mixed) and two with the full architecture fine-trained (BB-Fit3D and BB-Mixed). We evaluated each model separately on the test sets of Fit3D and BlanketGen2-Fit3D to assess their performance across both datasets.

\subsubsection{Evaluation on SLP}
\label{sec:slp_map}

To evaluate the effectiveness of the proposed synthetic dataset in enhancing the HPE architecture, we tested the trained models on the SLP dataset to assess the transferability of synthetic data-driven training to real-world, blanket-occluded scenarios. The SLP dataset includes each in-bed pose both without a blanket and with two different blanket occlusions
To measure the performance of the models with real blankets we tested separately the images with and without blankets (SLP-cover, SLP-uncover).

\begin{figure}[htpb]
\centering
\begin{subfigure}{.5\linewidth}
  \centering
  \includegraphics[height=8cm]{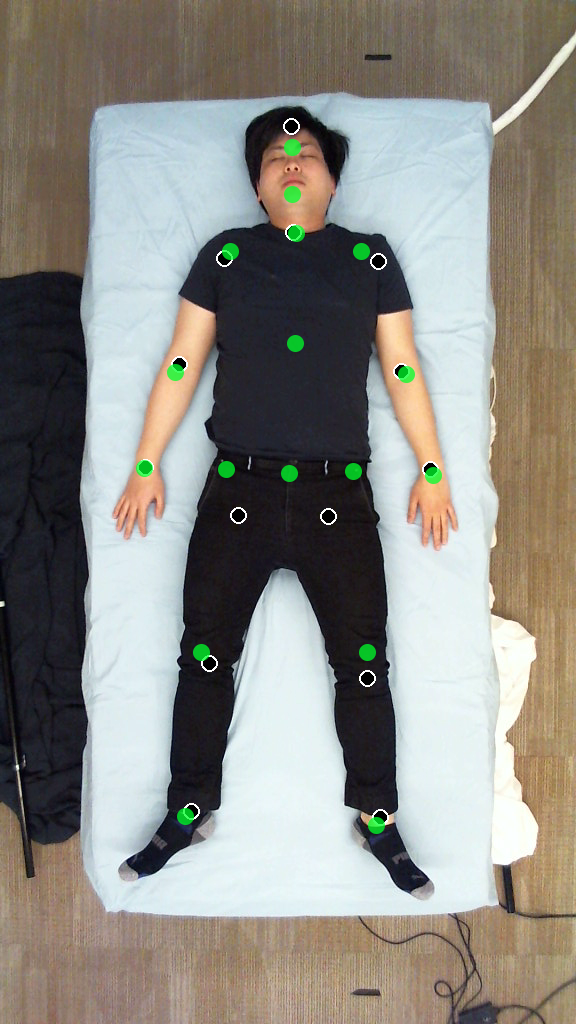}
  \caption{Frame of SLP}
  \label{fig:jointsslp}
\end{subfigure}%
\begin{subfigure}{.5\linewidth}
  \centering
  \includegraphics[height=8cm]{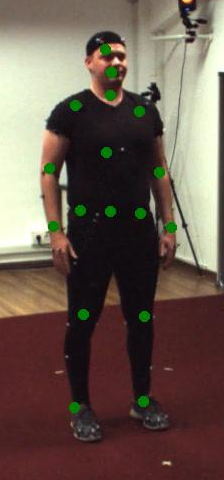}
  \caption{Frame of Fit3D}
  \label{fig:jointsfit3d}
\end{subfigure}
\caption{(a) A frame from SLP, the black and white circles are the ground-truth joints and the green circles are the joint predictions from the model with the VitPose-B model with the head fine-tuned on Fit3D (FT-Fit3D). Note how the hip joints are offset: in the SLP format, they are annotated on the axis of rotation of the leg, while in the Fit3D format, they are annotated at the outside of the hip. (b): A zoomed-in frame from Fit3D with the ground-truth joints in green.}
\label{fig:test}
\end{figure}

SLP utilizes a 14-joint skeleton format defined in \cite{johnson2010clustered}, which differs both from Fit3D and MS-COCO. Nevertheless, for every SLP joint, there is a similar Fit3D joint so it is possible to match them. To evaluate the models we mapped the predicted
Fit3D joints to the SLP joints. However, despite the correspondence between the joints, they are not precisely annotated in the same locations. For instance, the hip joints in SLP are positioned lower on the hips than in Fit3D, introducing a consistent error for all of our evaluations on SLP. This is illustrated in Figure \ref{fig:jointsslp} showing the two joint formats on the same frame from SLP, while Figure \ref{fig:jointsfit3d} displays the Fit3D joints on a frame from Fit3D. 

This discrepancy does not substantially affect intra-dataset architectural comparisons. While absolute evaluations are impacted, the relative performance of different architectures within the same dataset remains both valid and insightful for comparative analysis.

\subsubsection{Metrics utilized}
\label{sec:metrics}

We evaluated these models using standard 2D HPE metrics. PCK (Percentage of Correct Keypoints) measures the percentage of keypoints within a specified distance from the ground truth; for this metric, we used a threshold of 5\% of the bounding box size (PCK@0.05). NME (Normalized Mean Error) quantifies the average joint position error normalized by the distance between the head and thorax. NME is particularly useful for comparing results across images with varying resolutions and between datasets. This metric is crucial to understand the baseline error introduced by differences in joint definitions between Fit3D and SLP, by providing insight into the normalized error on the datasets, without being masked by PCK's baseline error threshold. 

\section{Results}

\begin{figure*}[hb]
    \includegraphics[width=0.95\textwidth]{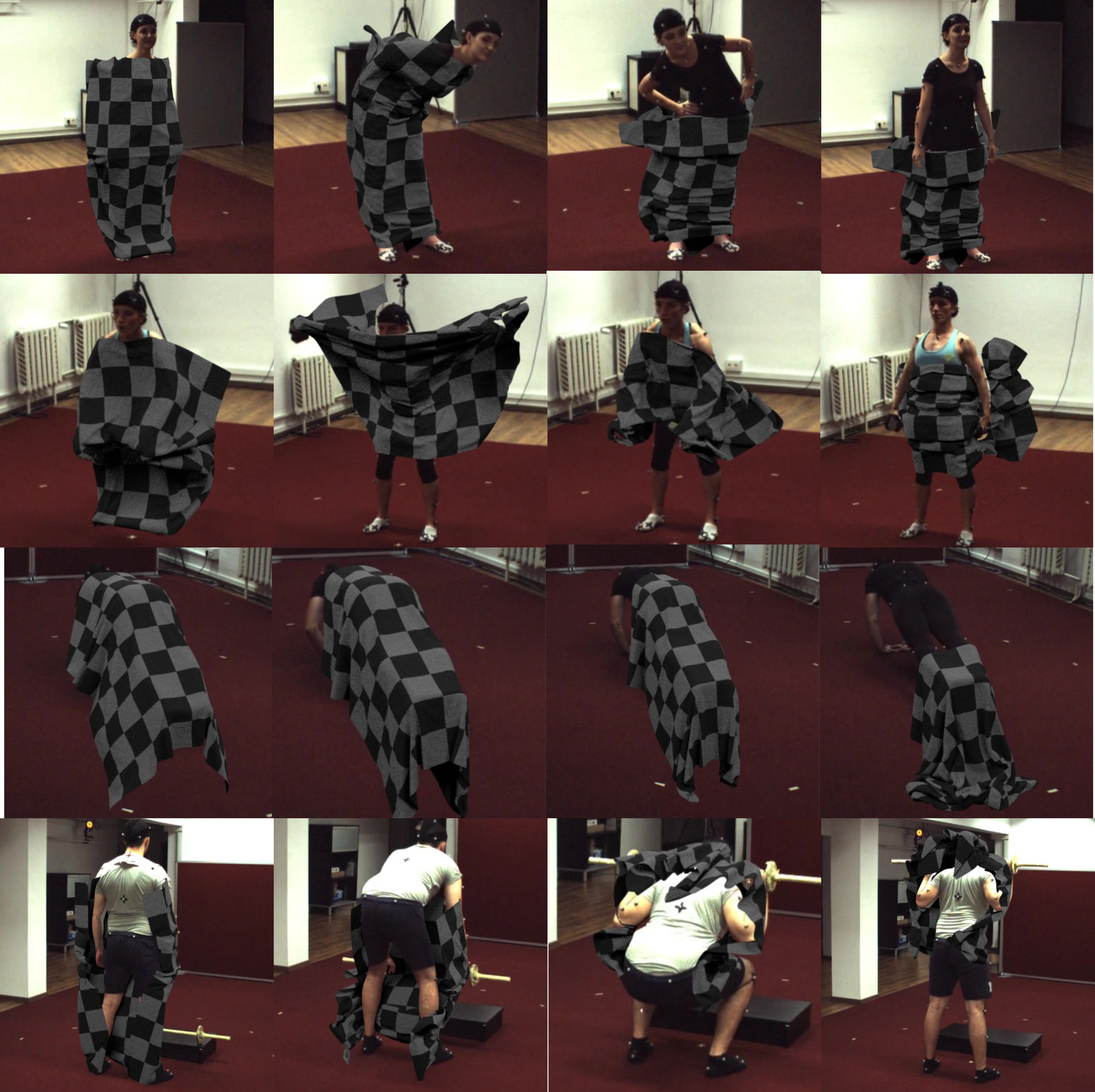}
    \caption{Four sequences from BlanketGen2-Fit3D, representing the dynamics of the simulated blankets during distinct movements, presented from the four different camera perspectives. From top to bottom, the sequences are named "warmup\_12", "side\_lateral\_raise", "diamond\_pushup", "squat"}
    \label{fig:frames}
\end{figure*}

\subsection{Dataset}
\label{sec:resutls_dataset}

The generated BlanketGen2-Fit3D dataset consists of 8 subjects performing 38 movement sequences, originating from 304 Fit3D videos. Some sequences required multiple sequential blanket simulations due to blankets falling off and needing resets, leading to a total of 655 unique simulations. The dataset consists of 1,214,812 frames in total, with a resolution of 900x900 pixels for the rendered images (Table \ref{tab:blanketgen3_fit3d}). Additionally, the .blend files for all blanket simulations are included, enabling the generation of as many new frames with different blanket textures and render settings as needed.

Figure \ref{fig:frames} presents selected frames from the generated dataset, with each row corresponding to a single repetition of a distinct movement captured from a unique camera angle. The dynamic deformation of the blanket is illustrated in response to the varying movement sequences.

\begin{figure}[htpb]
\centering
\includegraphics[width=1\linewidth]{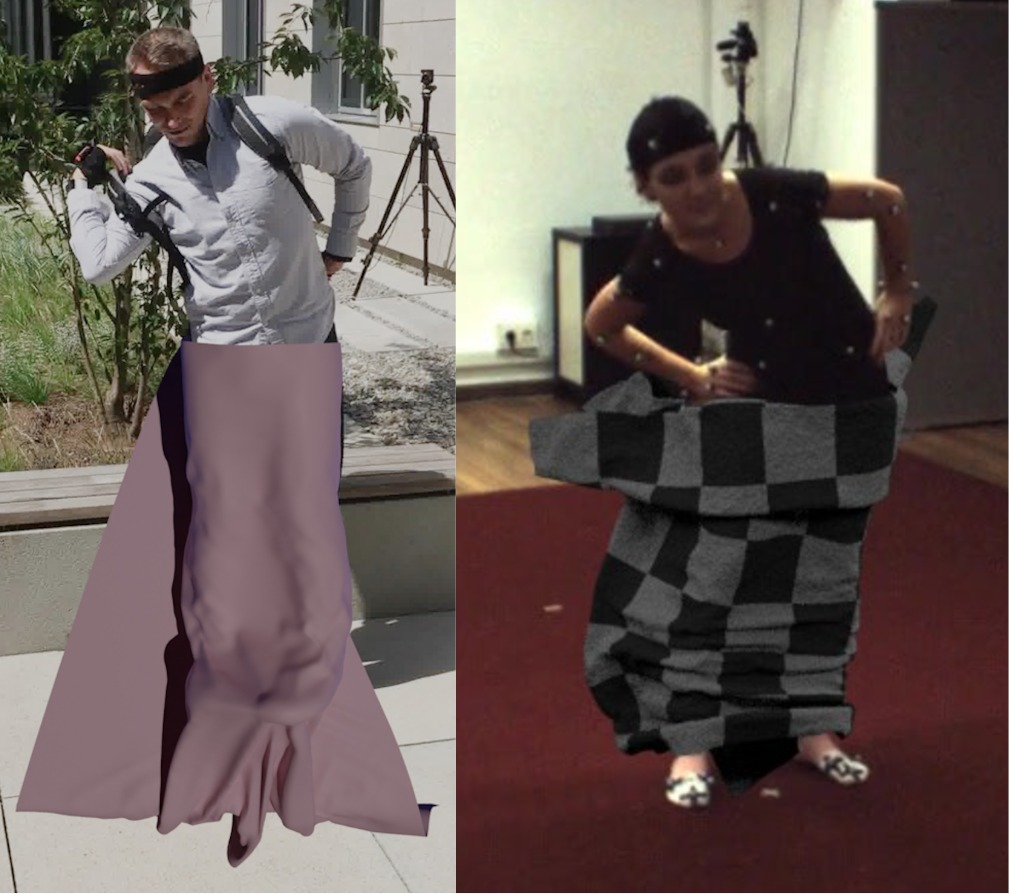} 
\caption{Comparing synthetic blanket occluded frames from BlanketGen-3DPW \cite{blanketgen} (left) and BlanketGen2-Fit3D (right) presented in this paper, showcasing the main advantages and improvements introduced by utilizing the new dataset, lighting setup, and texture.}
\label{fig:3dpw}
\end{figure}

In Figure \ref{fig:3dpw} we compare the blankets generated with the previous state of the art \cite{blanketgen}, and showcase the main advantages introduced by the new pipeline and the new data set. In the BlanketGen2 pipeline, the realism of the generated blanket is enhanced due to the consistent and static lighting of Fit3D, which allows for more accurate replication. The shadows are significantly softer, and the wrinkles of the blanket naturally cast shadows beneath them, aligning with the expected physical behavior. In contrast, the earlier setup had harsh shadows and was misaligned with the lighting in the scene. This inaccuracy stemmed from the use of a sun-like light emitting parallel rays, a choice intended to mimic real-world conditions in 3DPW. However, the pipeline's inability to account for the sun's direction resulted in shadows that appeared out of context and unrealistic.

The texture of woven fabric introduced in BlanketGen2 significantly enhances realism, but presents new challenges. By reducing the visibility of wrinkles, it becomes harder for models to interpret subtle details. In contrast, the previous state-of-the-art used a uniform blanket texture, making wrinkles more visible and providing clearer visual cues, although at the expense of realism. To balance this trade-off, BlanketGen2 incorporates a checkerboard color pattern of the woven texture, where the distortions of the squares reveal valuable information about surface deformations, even as the overall appearance becomes more complex to analyze.

\subsection{Trained Model Performance}
\label{sec:trained_model_performance}

\begin{table}[htbp]
\centering
\begin{tabular}{|l|c|l|ll|} 
\hline
Test Dataset & Blanket & Model & PCK$\uparrow$ & NME$\downarrow$ \\ 
\hline
\multirow{3}{*}{Fit3D} & \multirow{3}{*}{No} & FT-Fit3D & 0.983 & 0.147 \\
 &  & FT-Mixed & \textbf{0.984} & \textbf{0.142} \\
 &  & Difference & 0.001 & -0.005 \\ 
\hline
\multirow{3}{*}{BG2-Fit3D} & \multirow{3}{*}{Synthetic} & FT-Fit3D & 0.933 & 0.230 \\
 &  & FT-Mixed & \textbf{0.977} & \textbf{0.149} \\
 &  & Difference & 0.044 & -0.081 \\ 
\hline
\multirow{3}{*}{SLP-uncover} & \multirow{3}{*}{No} & FT-Fit3D & \textbf{0.810} & \textbf{0.262} \\
 &  & FT-Mixed & 0.798 & 0.279 \\
 &  & Difference & -0.012 & 0.017 \\ 
\hline
\multirow{3}{*}{SLP-cover} & \multirow{3}{*}{Real} & FT-Fit3D & 0.313 & 1.179 \\
 &  & FT-Mixed & \textbf{0.336} & \textbf{1.115} \\
 &  & Difference & 0.023 & -0.064 \\
\hline
\end{tabular}
\caption{The results obtained with each model on the different test sets. The numbers in bold are the best results for the test set. FT-Fit3D refers to the model that was fine-tuned with Fit3D, and FT-Mixed to the one fine-tuned on the mixed dataset. BG2-Fit3D is an abbreviation of BlanketGen2-Fit3D and SLP-cover and SLP-uncover are the SLP sets with and without blanket occlusions respectively. The PCK, and NME metrics are explained in section \ref{sec:metrics}.}
\label{tab:results}
\end{table}

Table \ref{tab:results} compares the performance of the two architectures with the trained deconvolutional heads and with the frozen backbones, one trained solely on the Fit3D dataset (FT-Fit3D) and one trained on a mixed dataset (FT-Mixed), across the four distinct test sets.

Under blanket occlusion conditions, the FT-Mixed model outperforms FT-Fit3D on both synthetic (BG2-Fit3D) and real (SLP-cover) blanket scenarios, achieving consistently higher PCK and lower NME values. While both architectures' performance naturally drops compared to the no occlusion scenarios.
Without occlusions FT-Mixed shows even a slight improvement on Fit3D, with the only exception on the SLP-uncover test set, where FT-Fit3D achieves marginally better results.

To interpret the results on SLP, it should be kept in mind that two factors contribute to the generally lower performance. First, the slightly different joint format as discussed in section \ref{sec:slp_map}, therefore the absolute performance evaluations are negatively impacted, nevertheless intra-dataset evaluations, comparing the performance of the two architectures on the same test set, still provide valuable insights. Secondly, there is a distribution shift of the data, from people mostly standing to lying in the bed.

\begin{table}[htb]
\centering
\begin{tabular}{|l|c|l|ll|} 
\hline
Test Dataset & Blanket & Model & PCK$\uparrow$ & NME$\downarrow$ \\ 
\hline
\multirow{3}{*}{Fit3D} & \multirow{3}{*}{No} & BB-Fit3D & \textbf{0.993} & 0.121 \\
 &  & BB-Mixed & 0.992 & \textbf{0.118} \\
 &  & Difference & -0.001 & -0.003 \\ 
\hline
\multirow{3}{*}{BG2-Fit3D} & \multirow{3}{*}{Synthetic} & BB-Fit3D & 0.935 & 0.208 \\
 &  & BB-Mixed & \textbf{0.990} & \textbf{0.116} \\
 &  & Difference & 0.055 & -0.092 \\ 
\hline
\multirow{3}{*}{SLP-uncover} & \multirow{3}{*}{No} & BB-Fit3D & \textbf{0.791} & \textbf{0.293} \\
 &  & BB-Mixed & 0.751 & 0.337 \\
 &  & Difference & -0.012 & 0.044 \\ 
\hline
\multirow{3}{*}{SLP-cover} & \multirow{3}{*}{Real} & BB-Fit3D & 0.252 & 1.399 \\
 &  & BB-Mixed & \textbf{0.265} & \textbf{1.398} \\
 &  & Difference & 0.013 & -0.001 \\
\hline
\end{tabular}
\caption{Results of the models with fine-tuned backbones and heads. The numbers in bold are the best results for the test set. BB-Fit3D refers to the model that was fine-tuned with Fit3D, and BB-Mixed to the one fine-tuned on the mixed dataset. BG2-Fit3D is an abbreviation of BlanketGen2-Fit3D and SLP-cover and SLP-uncover are the SLP sets with and without blanket occlusions respectively. The PCK, and NME metrics are explained in section \ref{sec:metrics}.}
\label{tab:resultsbackbone}
\end{table}

\subsubsection{Backbones finetuned}

The results of the additional fine-tuning, achieved by unfreezing the backbone layers and employing a reduced learning rate, are presented in Table \ref{tab:resultsbackbone}. BB-Fit3D refers to the model that was fine-tuned with Fit3D, and BB-Mixed to the one fine-tuned on the mixed dataset.

The trend of these results is similar to the ones with just the heads fine-tuned. Here, both models have slightly higher performance on Fit3D and BlanketGen2-Fit3D, however, they provide worse performance on SLP. These results indicate that by unlocking the backbones, the models may start forgetting the knowledge acquired from the greater diversity in the distribution of the original training datasets, and start overfitting to the distribution of Fit3D.

\section{Discussion}
\subsection{Synthetic dataset generation}

Acquiring real-world data with blanket occlusion and ground truth joint locations to train HPE architectures for in-bed HPE estimation is extremely challenging, as demonstrated by the authors of the SLP dataset \cite{slp}, which is currently the only available RGB blanket occluded HPE dataset, to the best of our knowledge. Despite the huge effort to acquire these datasets, it is relatively small and only able to present static scenes, as the blankets could only be placed on top of the static subjects, stemming from the ground truth labeling method, neglecting dynamic deformation patterns of the blankets. 

To train a robust HPE architecture, for an eventual real-world application, in this scenario, it is essential to utilize dynamic scenes and blanket deformations during training, to represent the expected distribution in the real world. Here, the utilized dynamic HPE dataset (Fit3D), and generated dynamic simulations of the blankets based on the ground truth dense human meshes, are composed into dynamic scenes aiming to address this. The simulated blankets realistically mimic how a fabric interacts with the body, generating natural wrinkles, wrapping, folding, and falling under gravity, while also responding dynamically to movements such as being flung or waved, closely adhering to the subjects' motions (Figure \ref{fig:frames}). Moreover, the texture and lighting utilized for these simulations provide a realistic render of the blankets, as we discussed and compared to the previous state of the art in Sec. \ref{sec:resutls_dataset}.

Our method offers a cost-effective dataset generation for this domain with ground truth HPE labels. Here we generated the largest dataset available, significantly larger compared to the previous state-of-the-art BlanketGen-3DPW or the real-world dataset SLP; BlanketGen2-Fit3D is 12.5X and 82.5X larger, respectively. Nevertheless, this dataset could be further extended, as the shared ".blend" files of the dynamic simulations enable virtually unlimited new renders with customizable textures. Moreover, the simulation parameters can be adjusted to model various blanket behaviors, including variations in material properties or thickness. Additionally, this method is applicable to any 3D HPE datasets with dense human mesh model ground truth (e.g SMPL-X), allowing the further extension of the dataset to encompass an even greater diversity of poses.

A limitation of the method is the domain shift of movements, as their distribution and representation may differ from typical in-bed movements. Nevertheless, the generated interaction with the fabric and the occlusions remain transferable, covering a wide variety of interactions.
While the utilized indoor dataset (Fit3D) has many advantages, it does limit the diversity of movements to a certain degree, which can be beneficial by excluding some actions, such as walking, that can easily throw the blankets off, requiring frequent resets in the simulation. The utilized four viewpoints improve viewpoint invariance, although the rear viewpoint often provides less detail and is less affected by blanket occlusion.

In summary, even though the blankets are synthetic, the combination of the physics-based cloth simulations, photorealistic texture, and indoor lighting setup make them look realistic and provide the largest HPE dataset for in-bed blanket occluded HPE.

\subsection{Test Results}

The four trained architectures demonstrated excellent performance on the Fit3D test set, representing the successful transfer learning from the COCO skeleton format to the Human3.6m skeleton format of Fit3D (PCK: 0.983-0.993; NME: 0.118-0.147; Table \ref{tab:results}, and \ref{tab:resultsbackbone}). These results (Table \ref{tab:results}) also show the feasibility to transfer the COCO pre-training of the ViTPose architecture, by only fine-tuning the heads (FT-Fit3D, FT-Mixed) to accurately estimate joints in the Human3.6M format. Furthermore, training with the mixed dataset has similar performance on the unoccluded test set ($\pm$0.001 PCK, NME: -0.003-0.005), indicating that incorporating synthetic blankets neither significantly enhances nor degrades performance in scenarios tested without blankets.

When comparing the architectures' performance on the BlanketGen2-Fit3D test set, the model utilizing the mixed dataset training (FT-Mixed) had a significantly higher accuracy (PCK: 0.977) compared to the model trained only on non-occluded data (FT-Fit3D, PCK: 0.933), with a difference of 0.044 PCK. This result of the FT-Mixed architecture demonstrates the potential to learn visual features of the deformations produced by the dynamic body-blanket interactions, improving the generalized estimation of human poses occluded by synthetic blankets. Additionally, compared to the baseline performance on the non-occluded Fit3D test set, the FT-Mixed architecture demonstrates strong robustness, maintaining similar performance even when synthetic blankets are present with only a 0.007 PCK decrease (0.984$\rightarrow$0.977). In contrast, the FT-Fit3D architecture is noticeably more impacted, with a 0.05 PCK reduction (PCK: 0.983$\rightarrow$0.933). These results show that the FT-Mixed architecture performs and generalizes well in both synthetic blanket occluded and non-occluded scenarios.

Evaluating the performance on the SLP test sets (uncovered and covered), we have to keep in mind the significant impact of the different joint formats discussed previously (Section \ref{sec:slp_map}, and \ref{sec:trained_model_performance}). This affected both models and FT-Fit3D slightly outperformed FT-Mixed on SLP without covers (FT-Fit3D: 0.810 PCK, FT-Mixed: 0.798 PCK, difference: -0.012 PCK). This PCK number without covers, in this case, is still considered good, as it is lower mainly due to the different joint format. Nevertheless, the two architectures are still performing comparable on the SLP-uncover test set, similar to how they performed on Fit3D. These results serve as a baseline for comparing the impact of real blanket occlusions. 
The performance of both models was significantly more impacted by the real blankets, than it was with synthetic blanket occlusions, in case of FT-Mixed: -0.462 PCK and -0.007 PCK respectively. This points to a substantial distribution shift remaining between the BlanketGen2-Fit3D and SLP datasets, which we discuss later in detail. Nonetheless, the FT-Mixed model outperformed FT-Fit3D in real-world scenarios with real blanket occlusions (FT-Fit3D: 0.313 PCK, FT-Mixed: 0.336 PCK, difference: 0.023 PCK). To further evaluate the impact of synthetic blanket occlusion training, NME may provide a more informative measure, though PCK was also improved. Introducing synthetic blanket occlusions during training improved the NME of the FT-Mixed architecture compared to FT-Fit3D by 0.081 on the BG2-Fit3D test set and 0.064 on the SLP Cover test set, demonstrating its positive impact both on synthetic and real blanket occlusions. These results transferred to real blankets are very promising. It shows that the training with synthetic blanket augmentations generated with BlanketGen2 on a dataset like Fit3D, where people perform activities mostly standing up, may be transferred to a real-world impact on a blanket occluded in-bed dataset like SLP, bridging the domain gap and distribution shift. However, substantial future work is necessary to realize the full potential of the methods introduced here, which we detail below in Section \ref{sec:future}.

The main distribution shift introduced by SLP originates from the ground-truth labeling method utilized, as these scenes had to be static and not dynamic. Without the dynamic interaction of the body and the blankets, there are significantly fewer wrinkles and visual features to be utilized for pose estimation. Additionally, these wrinkles are further masked by the lack of any patterns on the blankets and in some images the camera parameters are not perfectly calibrated, causing the people to be out of focus, or over- or under-exposed, further masking useful visual features. These effects compound to increase the difficulty of pose estimation on SLP, without reflecting real-world in-bed dynamic conditions.

The fine-training with the unlocked backbones improved the performance on Fit3D and BG2-Fit3D even more. However, with this better fit to these datasets some generalization was sacrificed, demonstrated by their performance on SLP, additionally losing some occlusion robustness from the MAE pre-training of VitPose, due to the restricted dataset size and training setup. This points out the future works for the training strategy, and datasets, which we detail in the next section, while this work already demonstrated the transferability of dynamic synthetic blanket occlusions to real-world datasets.

\subsection{Future Work}
\label{sec:future}

In the future, naturally, more data could be used for training, to further improve the models' performance and generalization capabilities, including all the other ViTPose pre-training datasets, such as MS COCO \cite{mscoco} and AIC \cite{aic}, including other HPE datasets, such as Human3.6M \cite{h36m}, or in-bed HPE dataset like SLP \cite{slp}. However, it would require one to transfer the corresponding skeleton rigs (COCO, Human3.6M, SLP) into one common skeleton rig, which would further enable a more direct comaprision of performance.

Generating more synthetic data by expanding BlanketGen2-Fit3D with a greater variety of blankets, both in their physical properties as well as their textures, could further improve bridging the domain gap between real and synthetic domains. Since the .blend files with baked blanket simulations are provided in BlanketGen2-Fit3D, more variety in blanket textures could be produced by including the rendering of the blanket as an even an online augmentation step in the model's training pipeline with the texture parameters being randomly sampled from a predetermined distribution.

Future research directions could include further explicit occlusion aware training for example MAE pretaining of blanket occluded scenes and contrastive learning approaches, since for every frame in BlanketGen2-Fit3D there is another frame in Fit3D with identical ground truth, these could be used in pairs to help the model learn how blankets alter images and learn better the corresponding features.

Lastly, given the information embedded in the dynamic deformation of the blankets, future research could explore leveraging HPE architectures specifically designed to incorporate temporal information.

\section{Conclusions}
In conclusion, we introduced BlanketGen2 to simulate and render synthetic photorealistic blanket occlusion on top of any HPE dataset using SMPL ground truth. To ensure efficiency, and versatility it is carried out in two steps, cloth simulation and rendering, allowing it to take advantage of the different CPU and GPU usage of both of these steps. Additionally, the output of the simulation, .blend files, are also saved and shared enabling unlimited new texture and lighting conditions to be rendered based on them. 

We introduced BlanketGen2-Fit3D, a novel synthetic dataset for blanket-occluded HPE. Utilizing the BlanketGen2 pipeline we augmented the Fit3D dataset, comprising of 1,217,312 frames with synthetic blankets, and provided the corresponding .blend files to enable the generation of virtually unlimited frames with varying blanket textures and render parameters. To the best of our knowledge, this represents the largest synthetic blanket-occluded temporal dataset for dynamic "in-bed" scenarios. 

In order to evaluate the utility of our proposed dataset we fine-tuned and tested a ViTPose-B model in multiple settings,
utilizing the Fit3D and BlanketGen2-Fit3D datasets, and provided benchmark test results. We showed that the visual features of the dynamic deformation of the blankets are learned, demonstrated by the improved performance on the synthetic blanket occluded test set, when utilizing this data for training (FT-Fit3D: 0.933 PCK, FT-Mixed: 0.977 PCK, +0.044 PCK).

Additionally, we tested the fine-tuned models (FT-Fit3D FT-Mixed) on the real-world in-bed blanket occluded dataset SLP. On the SLP-cover test set, we demonstrated that improvements brought by fine-tuning on the mixed dataset (FT-Mixed) improved results with 0.023 PCK and -0.064 NME, when real blankets were present, showing the potential transferability of dynamic synthetic blanket occlusions to real-world datasets.

The results presented in this paper indicate that synthetic blanket augmentation is a useful tool to improve in-bed blanket occluded pose estimation from RGB images. In the future, more data and contrastive learning strategies could be utilized to further improve performance.
The BlanketGen2 code and generated blanket dataset will be made available to the public upon acceptance. 

\section{Acknowledgments}

This work was partially funded by Fundação para a Ciência e a Tecnologia under the scope of the CMU Portugal program Ref PRT/BD/152202/2021. DOI 10.54499/PRT/BD/152202/2021 (https://doi.org/10.54499/PRT/ BD/152202/2021). This work is financed by National Funds through the Portuguese funding agency, FCT - Fundação para a Ciência e a Tecnologia, within project LA/P/0063/2020. DOI 10.54499/LA/P/0063/2020 | https://doi.org/10.54499/LA/P/0063/2020

%
%
%
\bibliographystyle{IEEEtran}
\bibliography{main}

\section{Biography Section}

\vspace{-33pt}
\begin{IEEEbiography}[{\includegraphics[width=1in,height=1.25in,clip,keepaspectratio]{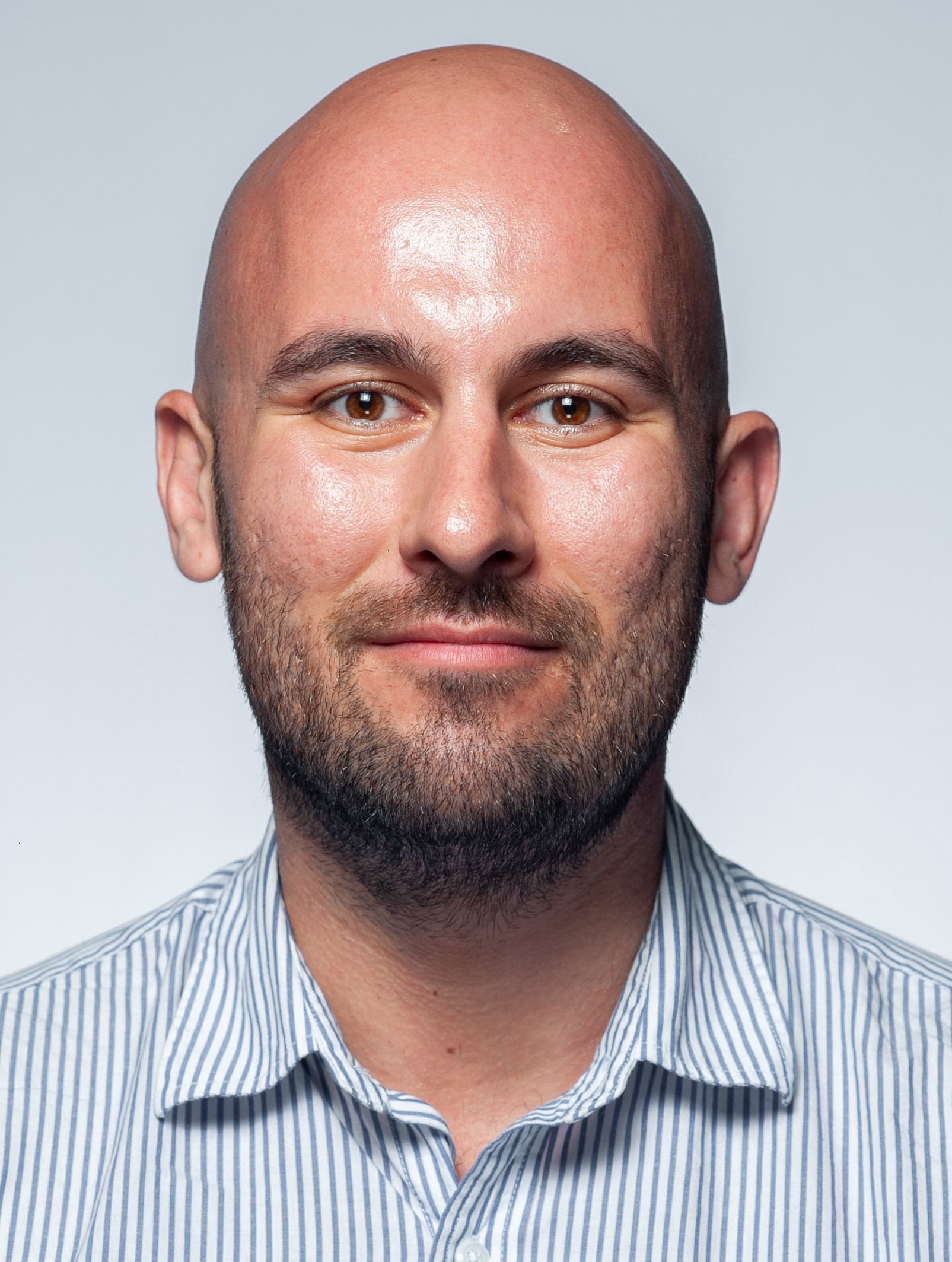}}]{Tamás Karácsony} is a Ph.D. candidate in the Carnegie Mellon Portugal affiliated Ph.D. (CMU Portugal) program in the Doctoral Program in Electrical and Computer Engineering (PDEEC), at the Department of Electrical and Computer Engineering of the Faculty of Engineering of the University of Porto (FEUP), Portugal, and a researcher at INESC-TEC: Institute for Systems and Computer Engineering, in the Center for Biomedical Engineering Research (C-BER) in the Biomedical Research And INnovation (BRAIN) research group. 

His PhD thesis "Explainable Deep Learning Based Epileptic Seizure Classification with Clinical 3D Motion Capture" is supervised by Prof. João Paulo Cunha and co-supervised by Prof. Fernando De la Torre. He is a visiting research scholar at the Computational Behavior (CUBE) Lab working with Prof. László A. Jeni, and at the Human Sensing Laboratory (HSL) working with Prof. Fernando De la Torre at The Robotics Institute (RI), Carnegie Mellon University (CMU). His research focuses on Advanced Human Sensing, 3D Motion Capture, Action and Pattern Recognition, Computer Vision, and Neuroengineering.

He earned an MSc degree with honours in Biomedical Engineering (2018) from the Technical University of Denmark (DTU), a BSc (2016), and an MSc with highest honours (2020) in Mechatronics Engineering from Budapest University of Technology and Economics (BUTE).
\end{IEEEbiography}

\vspace{-33pt}
\begin{IEEEbiography}[{\includegraphics[width=1in,height=1.25in,clip,keepaspectratio]{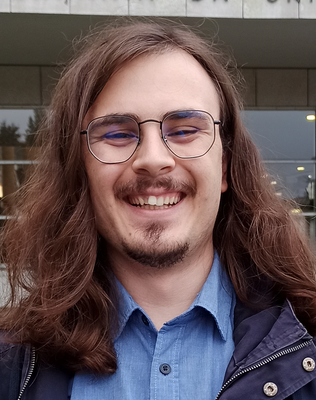}}]{João Carmona} received his BSc degree in Engineering Sciences (2020) as well as his MSc degree in Electrical and Computer Engineering (2022) from the Faculty of Engineering of the University of Porto (FEUP), Portugal. His master's thesis was supervised by Professor João Paulo Cunha, which he carried out at the Biomedical Research and INnovation (BRAIN) research group of the Center of Biomedical Engineering Research (C-BER) at INESC TEC: Institute for Systems and Computer Engineering, Technology and Science, where he also worked as a researcher for the following year (2023).
In his research, he explored the augmentation of Human Pose Estimation datasets with synthetic cloth occlusions as a potential approach to improve the performance of Deep Learning models in the context of patients in hospital beds.

\end{IEEEbiography}

\vfill 

\vspace{-33pt}
\begin{IEEEbiography}[{\includegraphics[width=1in,height=1.25in,clip,keepaspectratio]{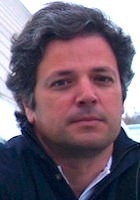}}]{João Paulo Cunha} is Full Professor of Bioengineering \& Electrical \& Computers Engineering (ECE) at the Department of ECE of the Faculty of Engineering of the University of Porto (FEUP), Portugal; Senior Researcher at the INESC-TEC: Institute for Systems and Computer Engineering where he created and coordinates the BRAIN – Biomedical Research And INnovation - research group and co-founded the Center for Biomedical Engineering Research (C-BER) that aggregates ~40 researchers. Prof. Cunha is also a mentor/co-founder and/or contributor to several MedTech/DeepTech startups (eight until now) by advising and licensing intellectual property of innovative biomedical technology developed jointly with his students for several years in his lab, such as Sword Health, iLof-Intelligent Lab-on-Fiber  and inSignals Neurotech. He is visiting professor at the Neurology Dep., Faculty of Medicine of the University of Munich, Germany since 2001 and at the Carnegie Mellon University – Silicon Valley Campus, USA, between 2016 and 2021. He serves as Scientific Director of the Carnegie-Mellon | Portugal program since 2014.

He earned a degree in Electronics and Telecommunications engineering (1989), a Ph.D. (1996) and an Habilitation (“Agregação”) degree (2009) in Electrical Engineering all at the University of Aveiro, Portugal.

Dr. Cunha is Senior Member of the IEEE – Institute of Electrical \& Electronics Engineers - (2004) – currently serving as Chair of the EMBS Portugal Chapter, member of the Editorial Board of NATURE/Scientific Reports and Associate Editor of FRONTIERS/Signal Processing journal. He is also habitual reviewer of several IEEE journals and other relevant scientific journals such as PLoS ONE or Movement Disorders. He is an internationally renowned expert in advanced biosignal processing, human motion analysis and neuroimaging. He has supervised and co-supervised more than 15 PhD \& Post-doc students in his areas of R\&D. He received several awards, being the most relevant the European Epilepsy Academy (EUREPA) “Best Contribution for Clinical Epileptology” Award in 2002. Prof. Cunha is co-author of +200 scientific publications and 43 patents from 10 patent families, holding an h-index of 35 (Google Scholar), with +5,000 citations.
\end{IEEEbiography}

\vfill

\end{document}